\documentclass[letterpaper]{article} % DO NOT CHANGE THIS
\usepackage{aaai2026}  % DO NOT CHANGE THIS
\usepackage{times}  % DO NOT CHANGE THIS
\usepackage{helvet}  % DO NOT CHANGE THIS
\usepackage{courier}  % DO NOT CHANGE THIS
\usepackage[hyphens]{url}  % DO NOT CHANGE THIS
\usepackage{graphicx} % DO NOT CHANGE THIS
\usepackage{natbib}  % DO NOT CHANGE THIS AND DO NOT ADD ANY OPTIONS TO IT
\usepackage{caption} % DO NOT CHANGE THIS AND DO NOT ADD ANY OPTIONS TO IT
\usepackage{booktabs} % For \toprule, \midrule, etc.
\usepackage{textgreek} % For Greek letters like μ
\usepackage[utf8]{inputenc}
\usepackage{newunicodechar}
\newunicodechar{₀}{$_0$}
\usepackage{newunicodechar}
\newunicodechar{₅}{$_5$}
\usepackage{tabularx}
\usepackage[utf8]{inputenc}
\usepackage{newunicodechar}

\usepackage{algorithm}
\usepackage{algorithmic}

\usepackage{newfloat}
\usepackage{listings}
\DeclareCaptionStyle{ruled}{labelfont=normalfont,labelsep=colon,strut=off} % DO NOT CHANGE THIS
\floatstyle{ruled}
\newfloat{listing}{tb}{lst}{}
\floatname{listing}{Listing}
\title{G2L: From Giga-Scale to Cancer-Specific Large-Scale Pathology Foundation Models via Knowledge Distillation}
\author{
    Yesung Cho\textsuperscript{\rm 1}, 
    Sungmin Lee\textsuperscript{\rm 2}, 
    Geongyu Lee\textsuperscript{\rm 1}, 
    Minkyung Lee\textsuperscript{\rm 2},
    Jongbae Park\textsuperscript{\rm 3},
    Dongmyung Shin\textsuperscript{\rm 1}\textsuperscript{\rm $\dagger$}
}
\affiliations{
    \textsuperscript{\rm 1}OmixAI Co. Ltd., Seoul, South Korea\\
    \textsuperscript{\rm 2}RadiSen Co. Ltd., Seoul, South Korea\\
    \textsuperscript{\rm 3}Kyunghee University, Seoul, South Korea\\
    \textsuperscript{\rm 1}\{yscho, gglee, shinsae11\}@omixai.com, 
    \textsuperscript{\rm 2}\{smlee, mklee317\}@radisentech.com, 
    \textsuperscript{\rm 3}jbp@khu.ac.kr\\
    \textsuperscript{\rm $\dagger$}Corresponding author
}

\usepackage{bibentry}
\begin{document}

\maketitle

\begin{abstract}
Recent studies in pathology foundation models have shown that scaling training data, diversifying cancer types, and increasing model size consistently improve their performance. However, giga-scale foundation models, which are trained on hundreds of thousands of slides covering tens of cancer types and contain billions of parameters, pose significant challenges for practical use due to their tremendous computational costs in both development and deployment. In this work, we present a novel strategy, named the G2L framework, to increase the performance of large-scale foundation models, which consist of only $15\%$ of the parameters of giga-scale models, to a comparable performance level of giga-scale models in cancer-specific tasks. Our approach applies knowledge distillation, transferring the capabilities of a giga-scale model to a large-scale model, using just 1K pathology slides of a target cancer (e.g., breast, prostate, etc.). The resulting distilled model not only outperformed state-of-the-art models of the same size (i.e., large-scale) across several benchmarks but also, interestingly, surpassed the giga-scale teacher and huge-scale models in some benchmarks. In addition, the distilled model exhibited a higher robustness index, indicating improved resilience to image variations originating from multiple institutions. These findings suggest that the proposed distillation approach for a large-scale model is a data- and parameter-efficient way to achieve giga-scale-level performance for cancer-specific applications without prohibitive computational burden. 
\end{abstract}


\section{Introduction}
Foundation models (FMs) have been increasingly adopted in computational pathology to extract morphological features from whole-slide images (WSIs). Especially, vision transformers (ViTs) trained with contrastive learning \cite{simclr, moco, dino, dinov2} or masked image modeling \cite{MAE, ibot} have achieved remarkable performance across various downstream applications, such as tumor classification \cite{BRCAS, breakHis} and gene mutation detection \cite{TP53}. 

Recent studies have shown that scaling the amount of training data, diversifying cancer types, and increasing model size  consistently improve the performance of FMs \cite{virchowv2, phikonv2}. For instance, giga-scale FMs, such as H-optimus-0 \cite{optimus} and GigaPath \cite{gigapath}, employ a ViT-Giga (ViT-G) backbone network with 1.9B parameters \cite{vit} and are trained on hundreds of thousands of pathology slides (e.g., 170K slides for GigaPath) that cover a wide range of cancer types (e.g., 28 types for GigaPath). Consequently, the gigascale models outperform their smaller-parameter counterparts, such as ViT-Huge (ViT-H) models with 0.6B and ViT-Large (ViT-L) models with 0.3B parameters, in a variety of downstream tasks.

\begin{figure}[t]
    \centering
    \includegraphics[width=0.47\textwidth]{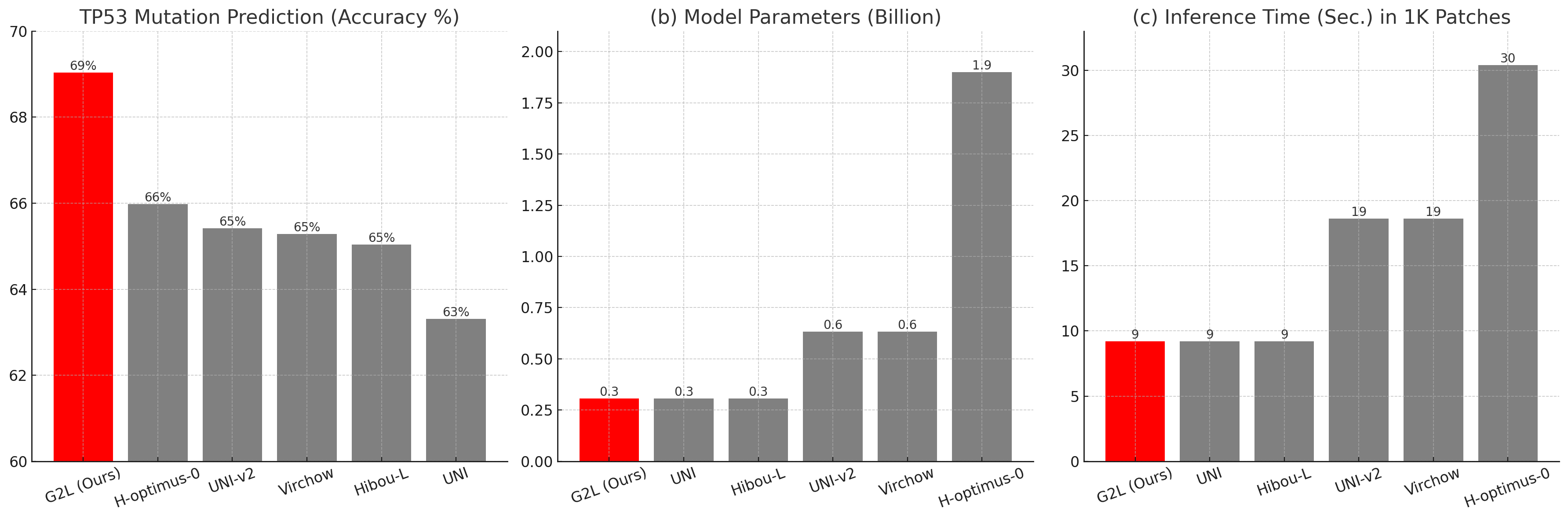}
    \caption{Comparison of G2L-distilled and other foundation models in TP53 mutation prediction benchmark measured by accuracy with model parameters (in billion) and inference times (in seconds). The G2L-optimized model outperforms the others in all metrics, as highlighted in red.}
    \label{fig:teaser}
\end{figure}

However, the computational demands of giga-scale FMs, which exceed a billion parameters and require hundreds of thousands of slides for training, present significant challenges for both model development and deployment. In particular, these resource-intensive requirements can limit accessibility in research institutions and clinical settings without a scalable computing infrastructure. In addition, while giga-scale FMs serve as a general feature extractor across various types of cancer, they can possibly dilute critical morphological signals specific to a certain cancer due to their significant heterogeneity \cite{picture, breat_unique}. For example, architectural patterns, nuclear morphology, and stromal interactions unique to breast carcinoma \cite{breat_unique} may be underrepresented when training data are dominated by other cancers. As a result, its performance can possibly be degraded for fine-grained discrimination within a specific cancer morphology.

To circumvent these challenges, we propose a novel giga-to-large (G2L) distillation framework to improve the performance of a large-scale FM to a comparable level of performance to a giga-scale FM via knowledge distillation. By G2L, a large-scale FM is optimized in a way to increase its sensitivity to a target cancer domain using only 1K pathology slides. 

In both breast and prostate cancers, we have evaluated the proposed G2L-distilled model against not only other large-scale FMs but also huge- and giga-scale counterparts across a wide range of downstream tasks, such as gene mutation prediction \cite{TP53}, tumor grade classification \cite{gleason}, and immune infiltration detection \cite{tils, tils2}. Our results showed that the distilled model outperformed existing large-scale FMs and even surpassed some huge- and giga-scale FMs in a set of benchmarks. These findings demonstrate that the proposed approach is a data- and parameter-efficient method to develop a high-performance, cancer-specific large-scale FM. Fig. \ref{fig:teaser} compares the G2L-distilled and other FMs based on the TP53 mutation prediction accuracy, model sizes, and inference times, showcasing its superiority over the others.

\begin{figure}[ht]
    \centering
    \includegraphics[width=0.5\textwidth]{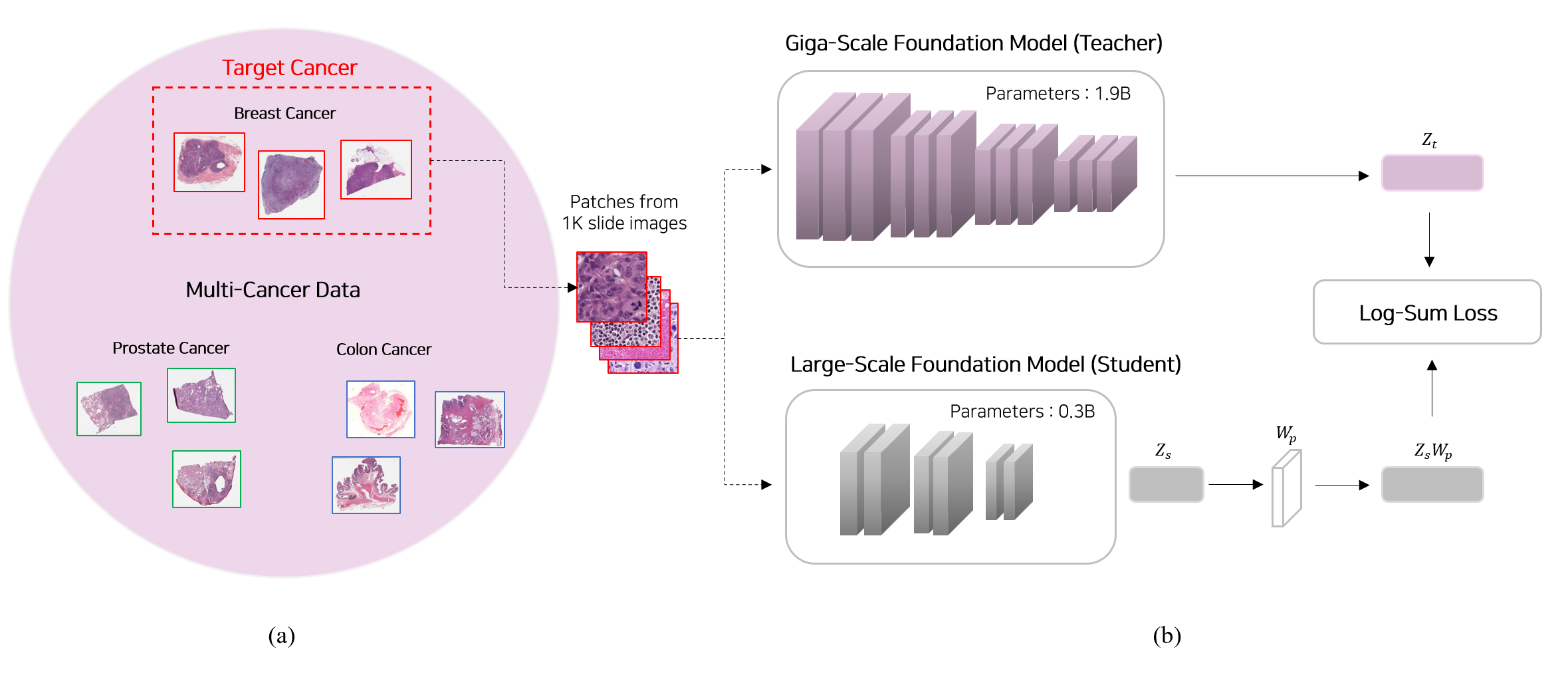}
    \caption{Overview of G2L framework.}
    \label{fig:G2L_framework}
\end{figure}

\section{G2L Framework}

\subsection{Overview}

Fig. \ref{fig:G2L_framework} illustrates the proposed G2L framework to build a cancer-specific, large-scale FM via knowledge distillation. First, a single cancer type that we want to target for the distillation (i.e., target cancer) is chosen in multi-cancer database (e.g., the cancer genome atlas (TCGA) \cite{TCGA}). Then, we select 1K pathology slides of the target cancer, where each of these slides is divided into a set of patches in the foreground tissue regions \cite{pathology_tile_sampling} (Fig. 2a). Finally, we distilled the knowledge of a giga-scale FM as a teacher to a large-scale FM as a student \cite{KD}  (Fig. 2b).

\subsection{Implementation Details}

For target cancers, we selected popular cancer types that are easily accessible in the public: breast and prostate. More precisely, we used non-overlapping patches with a size of 256 by 256 from 1K pathology slides (40x magnification) in TCGA-BRCA or TCGA-PRAD \cite{TCGA}, where each patch was randomly cropped to a size of 224 by 224 for training. We applied data augmentation to each patch before feeding it to teacher and student FMs, including horizontal/vertical flipping (50$\%$ chance), color jittering (50$\%$ chance; brightness = 0.15, contrast = 0.15, saturation = 0.1, and hue = 0.05), and Gaussian blurring (10$\%$ chance; kernel size = 9 x 9).

For giga- and large-scale FMs, H-optimus-0 \cite{optimus} (ViT-G/14 with 1.9B parameters), which has reported state-of-the-art performance in many downstream tasks in various cancers, was used as a teacher, and Hibou-L \cite{hibou} (ViT-L/14 with 0.3B parameters) was utilized as a student. Since these two FMs have different dimensions of output embeddings (1536 for H-optimus-0 and 1024 for Hibou-L), to match them with each other, we simply attached a linear projection layer with batch normalization at the end of the student network, following \cite{KD}. Finally, the loss function (i.e., Log-Sum Loss in Fig. \ref{fig:G2L_framework}) is defined as follows:

\begin{equation}
D(Z_s, Z_t; W_p) = \log \sum_i \left| Z_s W_p - Z_t \right|_{i}^{\alpha}
\end{equation}
where $Z_s$ and $Z_t$ are feature vectors of the student and teacher networks, respectively, $W_p$ is a linear project layer, and $\alpha$ is a smoothing factor (= 4).

\subsection{Hyperparameters and Computing Environment}

The training batch size was 32. We used an AdamW optimizer \cite{Adam} with an initial learning rate of 1×$10^{-4}$ and a weight decay of 0.05. Both the learning rate and the weight decay followed a cosine annealing schedule, decaying from 1×$10^{-4}$ to 1×$10^{-6}$ and from 0.05 to 0.5, respectively. We stopped distillation training when the current loss value exceeded the average loss value of the last 100 iterations more than 10 times. The computing environment was: three NVIDIA RTX A6000 GPUs, AMD Ryzen Threadripper PRO 3955WX CPU, and 506GB RAM.

\section{Experiments}

\subsection{Feature Similarity Before / After G2L}

We first investigated whether the student model learns a feature space similar to that of the teacher model after applying the G2L framework. To quantify the similarity, we employed centered kernel alignment (CKA) \cite{CKA}, which measures how similarly two models capture spatial information in their feature embeddings by computing the covariance between them and then normalizing the result to produce a similarity score between 0 and 1, where values closer to 1 indicate higher similarity.

To evaluate how the feature similarity changes before and after G2L, we used patches from two benchmark datasets, BRCAS and BreakHis (see details below). These patches were inferred by the teacher model (H-optimus-0) and the student model (Hibou-L), respectively. The resulting features were then used to compute CKA values, allowing us to assess the degree of spatial alignment in the latent space for each dataset.

\subsection{Downstream Task Benchmark }

To evaluate the performance of FMs as a general feature extractor without fine-tuning, we first adopted the non-training benchmark method similar to \cite{squeezing}. Concretely, we extracted 50-dimensional image feature vectors using principal component analysis and measured the performance based on the majority vote of the labels (e.g., tumor-infiltrated or not) of the features of k-nearest neighbors ($k$ = 15). For each downstream benchmark, a 10-fold cross-validation was performed. Here, the training data were not used to fine-tune the FMs but only to assign neighbor labels, thereby enabling a purely non-learnable evaluation of embedding quality.

To complement the non-parametric evaluation, we also performed a linear probing (LP) benchmark \cite{TITAN, SiMLP}, assessing the adaptability of the extracted representations of FMs for lightweight supervision. We trained a linear classifier on top of the frozen FM features using a 6:2:2 data split (train:validation:test). While the non-training method evaluates the intrinsic discriminative structure of features, this LP analysis verifies whether representations effectively support downstream tasks with minimal supervision.

For patch-level downstream benchmarks (e.g., TILS dataset below), each patch was fed into the FMs to extract image features, and the performance was measured accordingly. For slide-level and region-of-interest(ROI)-level downstream tasks (e.g., TP53 dataset below), non-overlapping patches from each slide or ROI were fed into the FMs to extract a set of features, which were then mean-pooled to generate an aggregated slide-level or ROI-level representation. 

All downstream benchmarks were either binary or multi-label classification tasks. In the non-training evaluation, we reported performance using accuracy. In the linear-probing benchmarks, we measured AUCs for binary tasks and one-vs-rest macro AUCs for multi-label classification.

For comparison, we have benchmarked not only the original teacher (H-optimus-0 ~\cite{optimus}) and student (Hibou-L ~\cite{hibou}) models but also other FMs with large- and huge-scales, including UNI (ViT-L with 307M parameters), UNI-v2 (ViT-H with 632M parameters) ~\cite{uni}, and Virchow (ViT-H with 632M parameters) ~\cite{virchow}.

\subsubsection{TILS (Tumor-Infiltrating Lymphocytes)} is a patch-level benchmark dataset extracted from the TCGA cohort \cite{tils}. It contains 5,245 patches of breast cancer, where each patch is annotated as binary labels (i.e., tumor-infiltrating lymphocytes or not).
 
\subsubsection{TP53} is a slide-level benchmark dataset derived from the TCGA breast cancer cohort  \cite{TP53}. We randomly sampled 511 slides out of a total of 1,133 slides for the benchmark, excluding any slides used for G2L. The binary label for each slide is TP53-mutation or not.

\subsubsection{IDC (Invasive Ductal Carcinoma)} is a patch-level benchmark dataset that comprises 162 breast cancer slides collected at Radboud University Medical Center \cite{IDC}. The total number of patches is 277,524, labeled as IDC-positive or IDC-negative.

\subsubsection{BRCAS (Breast Carcinoma Subtyping)} is an ROI-level benchmark dataset that contains 4,537 ROIs of 547 pathology slides from 189 breast cancer patients \cite{BRCAS}. The label for each ROI covers seven subtypes: normal, benign, usual ductal hyperplasia, flat epithelial atypia, atypical ductal hyperplasia, ductal carcinoma in situ, and invasive carcinoma.

\subsubsection{BreakHis} is an ROI-level benchmark dataset consisting of 7,909 ROIs from 82 slides (24 benign, 58 malignant) collected at the Universidade Federal do Paraná in Brazil \cite{breakHis}. The labels of benign ROIs include adenosis, fibroadenoma, phyllodes tumor, and tubular adenoma whereas the labels of malignant ROIs were ductal carcinoma, lobular carcinoma, mucinous carcinoma, and papillary carcinoma. The dataset provides each ROI image at four magnifications (40×, 100×, 200×, and 400×), enabling comprehensive evaluation across multiple fields of views. 

\subsubsection{Gleason} is a patch-level benchmark dataset that includes 77,364 patches of 331 tissue microarray (TMA) core images collected at the Vancouver Prostate Centre \cite{gleason}. For patch extraction, each TMA image was divided into non-overlapping patches with a size of 256 by 256, and each patch was assigned a label according to the majority vote of annotations from six pathologists. The labels for each patch comprise four categories: benign, Gleason grade 3, grade 4, and grade 5.

\subsubsection{AGGC} is a patch-level benchmark dataset that has 156,130 patches from 150 prostatectomy and 53 biopsy slides collected at Singapore's National University Hospital (NUH) \cite{aggc}. As similar as the Gleason benchmark, we assigned a label of each patch based on the majority class of annotations as five categories: benign, stroma, and Gleason grade 3, grade 4, and grade 5. 

\subsubsection{CHIMERA} is a slide-level benchmark dataset that incorporates 190 slides from 95 patients collected from several institutions, including Radboud University Medical Center \cite{chimera}. Each slide was labeled as biochemical recurrence positive or negative.

\begin{table}[b]
\centering
\scriptsize
\caption{Feature similarity measured by CKA values between teacher (H-optimus-0) and student (Hibou-L) models before and after applying G2L framework. The highest value of each row is highlighted in \textbf{bold}.}
\label{tab:feature_sim}
\begin{tabular}{l@{\hskip 6pt}cl}
\toprule
Dataset& CKA before G2L &CKA after G2L
\\
\midrule
BRCAS & 0.7594 ± 0.014  &\textbf{0.9683 ± 0.011} 
\\
BreakHis 40×& 0.8909 ± 0.003  &\textbf{0.9558 ± 0.001} 
\\
BreakHis 100×& 0.9147 ± 0.002  &\textbf{0.9686 ± 0.001} 
\\
BreakHis 200×& 0.9230 ± 0.001  &\textbf{0.9734 ± 0.000} 
\\
BreakHis 400×& 0.8995 ± 0.003  &\textbf{0.9575 ± 0.000} \\
\bottomrule
\end{tabular}
\end{table}

\begin{table*}[t]
\centering
\scriptsize
\caption{Benchmark results of existing foundation models of different sizes and the proposed G2L-distilled model. We evaluate accuracy in non-training method (\textbf{Top}) and AUC in linear probing (\textbf{Bottom}). The best performance is highlighted in \textbf{bold}, and the second-best is in \underline{underline}. *: evaluated based on binary labels (benign or malignant) and averaged over multiple magnifications. **: evaluated based on sub-typing labels (e.g., adenosis, fibroadenoma, etc.) and averaged over multiple magnifications.}
\begin{tabular}{lccccccc}
\toprule
Benchmark & Cancer & H-optimus-0 & UNI-v2 & Virchow & Hibou-L & UNI & G2L (Ours) \\
 & & (ViT-G; 1.9B) & (ViT-H; 0.6B) & (ViT-H; 0.6B) & (ViT-L; 0.3B) & (ViT-L; 0.3B) & (ViT-L; 0.3B) \\
\midrule
\multicolumn{8}{c}{\textit{Non-training Method (Accuracy)}} \\
\midrule
TILS & Breast & \underline{0.9344 ± 0.006} & 0.9291 ± 0.007 & 0.8900 ± 0.007 & 0.9214 ± 0.008 & 0.9334 ± 0.007 & \textbf{0.9362 ± 0.005} \\
TP53 & Breast & \underline{0.6598 ± 0.040} & 0.6542 ± 0.050 & 0.6528 ± 0.030 & 0.6504 ± 0.030 & 0.6331 ± 0.020 & \textbf{0.6904 ± 0.050} \\
IDC & Breast & 0.9141 ± 0.0003 & 0.9165 ± 0.0003 & 0.8813 ± 0.0004 & 0.9074 ± 0.0003 & \underline{0.9224 ± 0.0003} & \textbf{0.9232 ± 0.0001} \\
BRCAS & Breast & \underline{0.4526 ± 0.030} & \textbf{0.4638 ± 0.020} & 0.3677 ± 0.03 & 0.4103 ± 0.040 & 0.4305 ± 0.040 & 0.4520 ± 0.040 \\
BreakHis (Bin.)* & Breast & \textbf{0.8892} & 0.8742 & 0.8102 & 0.8393 & 0.8718 & \underline{0.8841} \\
BreakHis (Sub.)** & Breast & 0.5259 & \textbf{0.5502} & 0.4040 & 0.4660 & 0.5105 & \underline{0.5319} \\
Gleason & Prostate & \textbf{0.8994 ± 0.002} & 0.8678 ± 0.003 & 0.6162 ± 0.003 & 0.8124 ± 0.002 & 0.8798 ± 0.002 & \underline{0.8988 ± 0.001} \\
AGGC & Prostate & \underline{0.9226 ± 0.001} & 0.9170 ± 0.001 & 0.7540 ± 0.002 & 0.8788 ± 0.001 & 0.9194 ± 0.001 & \textbf{0.9243 ± 0.001} \\
CHIMERA & Prostate & \textbf{0.7663 ± 0.070} & 0.7605 ± 0.060 & 0.7368 ± 0.057 & 0.7184 ± 0.072 & 0.7394 ± 0.071 & \underline{0.7657 ± 0.050} \\
\midrule
\multicolumn{8}{c}{\textit{Linear Probing (AUC)}} \\
\midrule
TILS & Breast & 0.9822 ± 0.003 & 0.9788 ± 0.004 & 0.9819 ± 0.003 & \underline{0.9827 ± 0.002} & 0.9782 ± 0.004 & \textbf{0.9838 ± 0.002} \\
TP53 & Breast & \underline{0.7603 ± 0.052} & 0.6795 ± 0.054 & 0.7138 ± 0.070 & 0.7085 ± 0.076 & 0.7317 ± 0.066 & \textbf{0.8046 ± 0.048} \\
IDC & Breast & \underline{0.9778 ± 0.0008} & 0.9756 ± 0.001 & 0.9488 ± 0.002 & 0.9488 ± 0.002 & 0.9770 ± 0.0009 & \textbf{0.9796 ± 0.0009} \\
BRCAS & Breast & \underline{0.8226 ± 0.019} & \textbf{0.8489 ± 0.025} & 0.7943 ± 0.030 & 0.7809 ± 0.019 & 0.8146 ± 0.025 & 0.8156 ± 0.029 \\
BreakHis (Bin.)* & Breast & 0.9578 & \underline{0.9630} & 0.9471 & 0.9415 & \textbf{0.9657} & 0.9626 \\
BreakHis (Sub.)** & Breast & 0.8436 & \textbf{0.8711} & 0.7653 & 0.8033 & 0.8226 & \underline{0.8465} \\
Gleason & Prostate & \textbf{0.9846 ± 0.0008} & 0.9790 ± 0.001 & 0.9721 ± 0.001 & 0.9708 ± 0.001 & 0.9774 ± 0.001 & \underline{0.9841 ± 0.001} \\
AGGC & Prostate & 0.9955 ± 0.0001 & \underline{0.9956 ± 0.0002} & 0.9939 ± 0.0002 & 0.9921 ± 0.0002 & 0.9951 ± 0.0001 & \textbf{0.9958 ± 0.0002} \\
CHIMERA & Prostate & 0.8451 ± 0.090 & \textbf{0.8653 ± 0.106} & 0.7980 ± 0.092 & 0.8173 ± 0.081 & \underline{0.8572 ± 0.079} & 0.8520 ± 0.103 \\
\bottomrule
\end{tabular}\label{tab:2}
\end{table*}

\subsection{Robustness Index}
In addition, we quantified robustness of each model against image variations originating from multiple medical centers using the robustness index, following \cite{robustness}. We randomly sampled an equal number of patches (= 40) in each of five tissue classes (invasive tumor, in-situ tumor, tumor-associated stroma, inflamed stroma, and necrosis) spanning 21 medical centers in the TIGER dataset \cite{TIGER}. We first sampled a query patch, selected the top $k$ ($\in$ {3, 5, 10, 20}) nearest neighbor patches of the query, and measured the ratio of tissue consistency over center consistency (i.e., number of neighbor patches with the same tissue / number of neighbor patches with the same center). Therefore, larger robustness index ($>$ 1) represents better discrimination of biologically meaningful features than medical centers. We performed five-fold cross-validation for each FM.

\section{Results}

\subsection{Feature Similarity Before / After G2L}
Table \ref{tab:feature_sim} summarizes the feature similarity measured by CKA values between the teacher (H-optimus-0) and student (Hibou-L) models before and after applying the G2L framework. In the BRCAS benchmark, the CKA value was substantially increased after G2L (from 0.76 to 0.97), indicating that the student FM captures spatial information in the latent space highly similar to that of the teacher FM. A similar trend was observed in the BreakHis benchmark (e.g., from 0.89 to 0.96 in 40x magnification). Interestingly, the improvement was consistent across magnifications (40x, 100x, 200x, and 400x), suggesting that the G2L framework was successfully applied regardless of magnification scale.

\begin{table}[t]
\centering
\scriptsize
\caption{Robustness index of foundation models across different $k$ values. The best index is highlighted in \textbf{bold}.}
\label{tab:3}
\resizebox{\columnwidth}{!}{% % <--- 추가
\begin{tabular}{lcccc}
\toprule
Model & $k=3$ & $k=5$ & $k=10$ & $k=20$ \\
\midrule
H-optimus-0 & 1.0826 $\pm$ 0.03 & 1.1890 $\pm$ 0.04 & 1.3730 $\pm$ 0.07 & 1.8467 $\pm$ 0.12 \\
UNI-v2      & 1.0682 $\pm$ 0.03 & 1.2433 $\pm$ 0.05 & 1.4113 $\pm$ 0.05 & 1.8548 $\pm$ 0.10 \\
Virchow     & 1.0137 $\pm$ 0.04 & 1.1006 $\pm$ 0.03 & 1.1351 $\pm$ 0.04 & 1.2850 $\pm$ 0.07 \\
UNI         & 1.0701 $\pm$ 0.04 & 1.1460 $\pm$ 0.02 & 1.4110 $\pm$ 0.04 & 1.7811 $\pm$ 0.07 \\
Hibou-L     & 0.9056 $\pm$ 0.03 & 0.9905 $\pm$ 0.02 & 1.0879 $\pm$ 0.01 & 1.1855 $\pm$ 0.08 \\
\midrule
G2L (Ours)  & \textbf{1.0891 $\pm$ 0.02} & \textbf{1.3002 $\pm$ 0.10} & \textbf{1.5021 $\pm$ 0.06} & \textbf{2.0316 $\pm$ 0.09} \\
\bottomrule
\end{tabular}% % <--- 추가
} % <--- 추가
\end{table}

\subsection{Downstream Task Benchmark}
Table \ref{tab:2} compares the performance of existing FMs and the G2L-distilled FM across different benchmark tasks. In general, the G2L-distilled FM achieved top performance in many benchmarks (TILS, TP53, IDC, and AGGC) and ranked second in others (BreakHis and Gleason), with the exception of BRCAS. The original H-optimus-0, which is the largest giga-scale FM in this comparison, also achieved competitive results, ranking first in BreakHis-Binary and Gleason and second in TILS, TP53 and BRCAS. In addition, UNI-v2, which is a huge-scale FM, reported the best performance values in BRCAS and BreakHis-Subtyping.

The substantial performance gains of the G2L-distilled FM over the others has an important implication: Although the G2L model has only $15\%$ of the parameters of the giga-scale teacher FM (and $50\%$ parameters compared to the huge-scale models), it demonstrates better representational discrimination in multiple tasks, potentially capturing delicate, cancer-specific morphological features. For example, in the TP53 mutation prediction benchmark, the performance between H-optimus-0 and G2L (0.66 vs. 0.69) suggests that the G2L model can better discriminate subtle morphology-genomics correlations that may be underrepresented in the multi-cancer teacher. Similarly, the performance gap in the TILS detection benchmark, although modest, may indicate higher sensitivity of the G2L model to immune-related histological patterns.

\subsection{Robustness Index}
Table \ref{tab:3} presents the robustness index of different FMs across different number of nearest neighbors ($k$ = 3, 5, 10, and 20). The G2L-distilled model consistently surpassed both the teacher and all other models. These results support the clinical applicability of the G2L model, indicating its superior ability to prioritize biologically meaningful features over image variations originating from different institutions.

\section{Conclusion}

In this work, we demonstrated that the high performance of a giga-scale pathology FM can be achieved with a much smaller large-scale FM through efficient knowledge transfer. The proposed G2L framework significantly boosted model performance using only 1K pathology slides, reducing both data and computational demands while achieving giga-scale–level performance. Notably, the G2L-distilled model even outperformed its giga-scale teacher and larger counterparts in many downstream benchmarks. Overall, these findings highlight G2L as a practical and cost-effective method for developing powerful cancer-specific pathology FMs without prohibitive computational resources.

% These findings highlight a practical and cost-effective strategy for developing powerful cancer-specific pathology foundation models. 

% The results underscore the value of knowledge distillation in producing efficient, organ-specific foundation models without sacrificing performance. By tailoring the distilled model to breast cancer, we preserved critical morphological knowledge while improving feature sensitivity for downstream tasks. This aligns with our hypothesis that giga-scale, multi-cancer models may carry representational redundancies or noise when applied to single-cancer tasks.

 \newpage

\bibliography{aaai2026}


\end{document}